%% file: main.tex
\pdfoutput=1

\documentclass[11pt]{article}

\usepackage{emnlp2021}

\usepackage{times}
\usepackage{latexsym}

\usepackage[utf8]{inputenc}

\usepackage{microtype}

\input{preamble.tex}

\title{Compositional Networks Enable Systematic Generalization\\ for Grounded Language Understanding}

\author {
        Yen-Ling Kuo,
        Boris Katz,
        Andrei Barbu \\
    MIT CSAIL \& CBMM \\
    \{ylkuo, boris, abarbu\}@mit.edu
}

\begin{document}
\maketitle



\begin{abstract}
  Humans are remarkably flexible when understanding new sentences that include
  combinations of concepts they have never encountered before. Recent work has
  shown that while deep networks can mimic some human language abilities when
  presented with novel sentences, systematic variation uncovers the limitations
  in the language-understanding abilities of networks. We demonstrate that these
  limitations can be overcome by addressing the generalization challenges in the
  gSCAN dataset, which explicitly measures how well an agent is able to interpret
  novel linguistic commands grounded in vision, e.g., novel pairings of adjectives
  and nouns. The key principle we employ is compositionality: that the compositional
  structure of networks should reflect the compositional structure of the problem
  domain they address, while allowing other parameters to be learned end-to-end.
  We build a general-purpose mechanism that enables agents to generalize their
  language understanding to compositional domains. Crucially, our network has the
  same state-of-the-art performance as prior work while generalizing its knowledge
  when prior work does not. Our network also provides a level of interpretability
  that enables users to inspect what each part of networks learns.
  Robust grounded language understanding without dramatic failures and without
  corner cases is critical to building safe and fair robots; we demonstrate the
  significant role that compositionality can play in achieving that goal.
\end{abstract}



\section{Introduction}

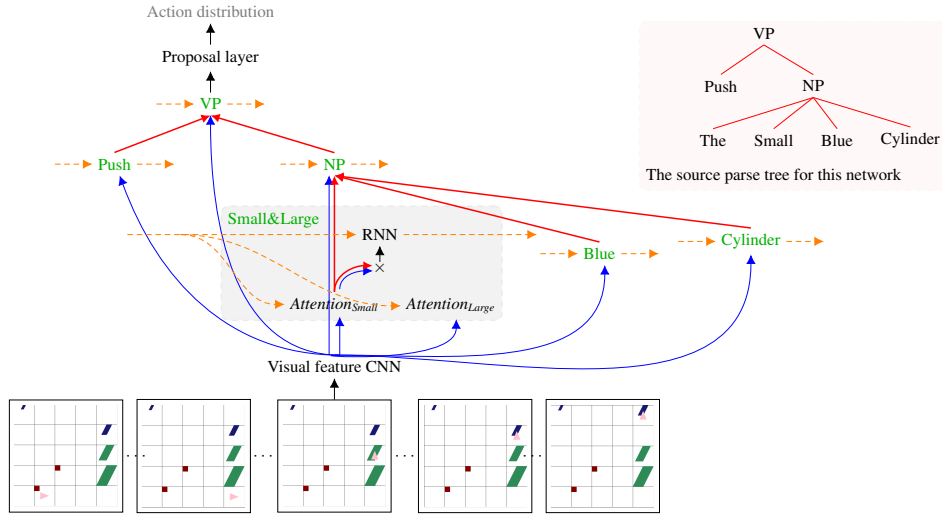
\begin{figure*}
  \centering
  \scalebox{0.7}{
  \begin{small}
    \begin{tikzpicture}
      \node[draw,solid,black] (frame) {\includegraphics[width=0.12\textwidth]{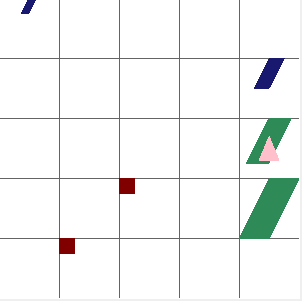}};
      \node[draw,solid,black,left=0.5 of frame] (frame1) {\includegraphics[width=0.12\textwidth]{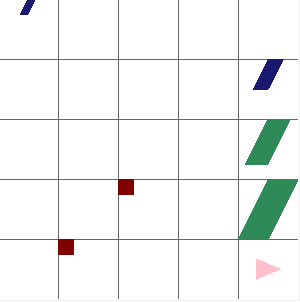}};
      \node[draw,solid,black,left=2.9 of frame] (frame2) {\includegraphics[width=0.12\textwidth]{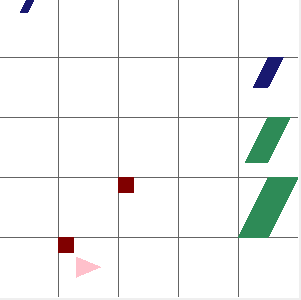}};
      \node[draw,solid,black,right=0.5 of frame] (frame4) {\includegraphics[width=0.12\textwidth]{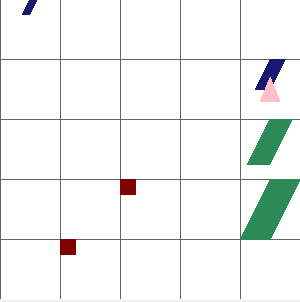}};
      \node[draw,solid,black,right=2.9 of frame] (frame5) {\includegraphics[width=0.12\textwidth]{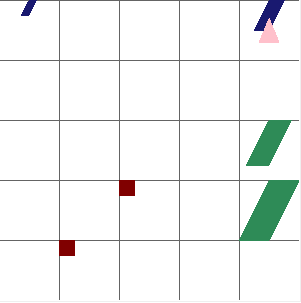}};
      \node[left=-0.09 of frame] () {$\cdots$};
      \node[left=2.3 of frame] () {$\cdots$};
      \node[right=-0.05 of frame] () {$\cdots$};
      \node[right=2.35 of frame] () {$\cdots$};
      \node[above=0.4 of frame] (encoder) {Visual feature CNN};
      \node[above=1.7 of encoder, xshift=35ex,Green] (blue) {Blue};
      \node[above=1.9 of encoder, xshift=55ex,Green] (cylinder) {Cylinder};
      \node[above=2.5 of encoder] (small) {};
      \node[below=0 of small, xshift=6ex] (RNN) {RNN};
      \node[below=0.2 of RNN] (x) {$\times$};
      \node[below=1.3 of small] (attention) {$\textit{Attention}_\textit{Small}$};
      \node[right=0.3 of attention, yshift=-0.2ex] (attention_big) {$\textit{Attention}_\textit{Large}$};
      \node[left=0.06 of small,Grey,Green,yshift=-0.4ex] (smallText) {Small\&Large};
      \node[above=3.4 of encoder,Green] (np) {NP};
      \node[left=3.4 of np,Green] (push) {Push};
      \node[left=1.7 of np, yshift=8ex,Green] (vp) {VP};
      \node[above=0.4 of vp] (proposal) {Proposal layer};
      \node[above=0.4 of proposal,Grey] (action) {Action distribution};
      \begin{scope}[on background layer]
        \node (corpus) [rounded corners, dashed, draw=black, fit= (small)(smallText)(attention)(attention_big)(RNN), fill=gray, opacity=0.1,
        inner sep=0.1ex] {};
      \end{scope}
      \begin{scope}[xshift=50ex,yshift=42ex]
      \node[Black] (pthe) {The};
      \node[right=0.3 of pthe,Black] (psmall) {Small};
      \node[right=0.15 of psmall,Black] (pmid) {};
      \node[right=0.3 of psmall,Black] (pblue) {Blue};
      \node[right=0.3 of pblue,Black] (pcylinder) {Cylinder};
      \node[above=0.7 of pmid,Black] (pnp) {NP};
      \node[left=1.0 of pnp,Black] (ppush) {Push};
      \node[left=0.3 of pnp, yshift=7ex,Black] (pvp) {VP};
      \draw[thin,Red] (pthe.north) -- (pnp.south);
      \draw[thin,Red] (pblue.north) -- (pnp.south);
      \draw[thin,Red] (psmall.north) -- (pnp.south);
      \draw[thin,Red] (pcylinder.north) -- (pnp.south);
      \draw[thin,Red] (pnp.north) -- (pvp.south);
      \draw[thin,Red] (ppush.north) -- (pvp.south);
      \node[below=0.2 of psmall] (plabel) {The source parse tree for this network};
      \end{scope}
      \begin{scope}[on background layer]
        \node (corpus) [rounded corners, dashed, draw=black, fit=(pthe)(pblue)(psmall)(pnp)(pcylinder)(plabel)(pvp), fill=red, opacity=0.03,
        inner sep=0.1ex] {};
      \end{scope}
      \draw[-{Latex[length=1.2ex,width=1.2ex]},thin,black] (frame) -- (encoder);
      \draw[-{Latex[length=1.2ex,width=1.2ex]},thin,Blue] (encoder.north) to[out=0,in=-90] (cylinder.south);
      \draw[-{Latex[length=1.2ex,width=1.2ex]},thin,Blue] ($(encoder.north)+(0.1,0)$) to[out=0,in=-90] ($(blue.south)+(0.1,0)$);
      \draw[-{Latex[length=1.2ex,width=1.2ex]},thin,Blue] ($(encoder.north)+(0.1,0)$) to[out=90,in=-90] ($(attention.south)+(0.1,0)$);
      \draw[-{Latex[length=1.2ex,width=1.2ex]},thin,Blue] ($(encoder.north)+(0.1,0)$) to[out=-180,in=-90] ($(attention_big.south)+(0.1,0)$);
      \draw[-{Latex[length=1.2ex,width=1.2ex]},thin,Blue] ($(encoder.north)+(-0.1,0)$) -- ($(np.south)+(-0.1,0)$);
      \draw[-{Latex[length=1.2ex,width=1.2ex]},thin,Blue] ($(encoder.north)+(0.1,0)$) to[out=180,in=-70] ($(push.south)+(0.1,0)$);
      \draw[-{Latex[length=1.2ex,width=1.2ex]},thin,Blue] ($(encoder.north)+(0.1,0)$) to[out=180,in=-90] ($(vp.south)+(0.0,0)$);
      \draw[-{Latex[length=1.2ex,width=1.2ex]},thin,Blue] ($(encoder.north)+(0.1,1.25)$) to[out=90,in=180] ($(x.west)+(0.1,-0.05)$);
      \draw[-{Latex[length=1.2ex,width=1.2ex]},thick,Red] (cylinder.north) -- (np.south);
      \draw[-{Latex[length=1.2ex,width=1.2ex]},thick,Red] (attention) -- (np);
      \draw[-{Latex[length=1.2ex,width=1.2ex]},thick,Red] (blue.north) -- (np.south);
      \draw[-{Latex[length=1.2ex,width=1.2ex]},thick,Red] (push.north) -- (vp.south);
      \draw[-{Latex[length=1.2ex,width=1.2ex]},thick,Red] (np.north) -- (vp.south);
      \draw[-{Latex[length=1.2ex,width=1.2ex]},thick,Red] (attention.north) to[out=90,in=180] ($(x.west)+(0.1,0.05)$);
      \draw[-{Latex[length=1.2ex,width=1.2ex]},thin,Black] (proposal) -- (action);
      \draw[-{Latex[length=1.2ex,width=1.2ex]},thin,Black] ($(vp.north)$) -- ($(proposal.south)$);
      \draw[-{Latex[length=1.2ex,width=1.2ex]},thin,black] ($(x.north)+(0,-0.1)$) -- (RNN);
      \draw[thin,orange,densely dashed,-{Latex[length=1.2ex,width=1.2ex]}] ($(push.west)+(-0.7,0)$) -- ($(push.west)+(0,0)$);
      \draw[thin,orange,densely dashed,-{Latex[length=1.2ex,width=1.2ex]}] ($(push.east)+(0,0)$) -- ($(push.east)+(0.7,0)$);
      \draw[thin,orange,densely dashed,-{Latex[length=1.2ex,width=1.2ex]}] ($(RNN.west)-(4.3,0)$) -- (RNN.west);
      \draw[thin,orange,densely dashed,-{Latex[length=1.2ex,width=1.2ex]}] (RNN.east) -- ($(RNN.east)+(2.5,0)$);
      \draw[thin,orange,densely dashed,-{Latex[length=1.2ex,width=1.2ex]}] ($(RNN.west)-(3.3,0)$) to[out=0,in=180] (attention.west);
      \draw[thin,orange,densely dashed,-{Latex[length=1.2ex,width=1.2ex]}] ($(RNN.west)-(3.3,0)$) to[out=0,in=180] (attention_big.west);
      \draw[thin,orange,densely dashed,-{Latex[length=1.2ex,width=1.2ex]}] ($(blue.west)+(-0.7,0)$) -- ($(blue.west)+(0,0)$);
      \draw[thin,orange,densely dashed,-{Latex[length=1.2ex,width=1.2ex]}] ($(blue.east)+(0,-0)$) -- ($(blue.east)+(0.7,-0)$);
      \draw[thin,orange,densely dashed,-{Latex[length=1.2ex,width=1.2ex]}] ($(cylinder.west)+(-0.7,0)$) -- ($(cylinder.west)+(0,-0)$);
      \draw[thin,orange,densely dashed,-{Latex[length=1.2ex,width=1.2ex]}] ($(cylinder.east)+(0,0)$) -- ($(cylinder.east)+(0.7,0)$);
      \draw[thin,orange,densely dashed,-{Latex[length=1.2ex,width=1.2ex]}] ($(np.west)+(-0.7,0)$) -- ($(np.west)+(0,0)$);
      \draw[thin,orange,densely dashed,-{Latex[length=1.2ex,width=1.2ex]}] ($(np.east)+(0,0)$) -- ($(np.east)+(0.7,0)$);
      \draw[thin,orange,densely dashed,-{Latex[length=1.2ex,width=1.2ex]}] ($(vp.west)+(-0.7,0)$) -- ($(vp.west)+(0,0)$);
      \draw[thin,orange,densely dashed,-{Latex[length=1.2ex,width=1.2ex]}] ($(vp.east)+(0,0)$) -- ($(vp.east)+(0.7,0)$);
    \end{tikzpicture}
  \end{small}
  }
  \caption{\small{The structure of the model interpreting and following \emph{Push
    the small blue cylinder}. In light red at the top right, we show the parse
    tree, as produced by a constituency parser. This tree is the source of the
    structure found within the compositional network; note the corresponding
    structure of the red lines. Each {\green token} in the parse becomes a
    recurrent network in the model, shown in green. Red lines show which
    recurrent networks are connected to one another through {\red attention maps}.
    Blue lines are {\blue visual observations}, available to every node. Orange
    lines are {\orange recurrent connections} allowing words to keep state. One module, 
    Small\&Large, is expanded, shown on a {\grey grey} background. This module 
    has two components which are trained to have opposite polarity. Each predicts 
    an attention map which then updates the hidden state of the word 
    and is passed to any subsequent words. The state of the root model
    is decoded into an action that the agent should execute next.}}
  \label{fig:model}
  \label{fig:rnn-model}
  \vspace{-2ex}
\end{figure*}

One of the defining characteristics of human languages is that they are
productive.
We can combine together concepts in novel ways to express ideas that have never
been thought of before.
This is for a good reason: as children, we observe very little of our world before
we must speak to others, meaning that even mundane language is novel and not
just parroting back something already expressed for us.
Similarly, even with massive data collection efforts, deep models can only have
an opportunity to observe a small subset of the possible utterances and worlds.
This problem becomes especially acute when those models must drive the behavior of a
robot, because misunderstanding a command may pose a serious safety hazard.

Recently, there have been a number of attempts to probe the understanding of deep
networks trained to perform linguistic tasks.
\citet{lake2018still} point out that generalization to novel compositions of
concepts is rather limited.
This is not a matter of the amount of data available; for example,
\citet{mccoy2019berts} find that even networks with the same test set performance
can have very different generalization abilities.
More recently, \citet{ruis2020gscan} released gSCAN for testing the generalization
abilities of grounded language understanding.
In gSCAN, an agent must follow a natural-language command in a 2D environment.
Commands of specific types are systematically held out; for example, no command
with a particular adjective-noun combination appears in the training set.
When the test set distribution is similar to the training set, performance is
phenomenal: 97\% of commands are executed correctly.
Yet, when combinations are missing from the training set, such as holding out an
adjective-noun pair like ``yellow squares'', only 24\% to 55\% of commands are
executed correctly.

Guided by the notion that compositionality is the central feature of human
languages which deep networks are failing to internalize, we construct a
compositional network to guide the behavior of agents.
Given a command, a command-specific network is assembled from previously-trained
modules.
Modules are automatically discovered in the training set without any annotation.
The network structure that combines those modules is derived from the linguistic
structure of the command.
In this way, the compositional structure of language is reflected in the
compositional structure of the computations executed by the network.

Compositionality is not specific to any one dataset -- it is a general principle --
and the implementation we provide here is not specific to gSCAN.
Even though our base network achieves the same 97\% performance in the random test set
as the state-of-the-art models for gSCAN, it generalizes significantly better in a
number of ways, including few-shot learning and longer action sequences.
Where this approach shines is predicted well by the types of compositionality
that exist in the network.
For example, novel combinations of concepts related to individual objects
perform well.
%
%
An additional benefit of compositional networks is that they open the door to
naturally including other linguistic principles.
For example, it appears that not all parses are made equal.
In our case, network structures derived from a semantic parser lead to better-performing
agents compared to structures derived from a constituency or dependency parser.
We show another example of this idea by incorporating the lexical semantics of
words, \eg antonyms, as an additional loss while training the network.
%

Our approach forgoes the most popular mechanism for increasing the
generalization performance of neural networks: data augmentation.
%
%
Data augmentation has substantial drawbacks: it is arbitrary, it slows down
training time, and it is dataset and problem specific.
In addition, data augmentation introduces many parameters that must be tuned and
much knowledge that must be provided by humans.
We show that the generic principle of compositionality can replace data augmentation
without any of these drawbacks.
%
It remains an open question whether every data augmentation approach has a
corresponding compositional structure that can supplant and generalize it.
Compositional approaches could be combined with data augmentation, potentially
raising their performance even further.

%


Our work makes four contributions.
\begin{compactenum}
\item We demonstrate a class of compositional networks which generalize the
  ability of agents to execute commands that contain novel combinations of concepts.
\item We systematically replace data augmentation with compositionality
  resulting in both higher performance and a simpler, principled, and dataset-agnostic
  method.
\item We incorporate the lexical semantics of words (\eg if they are antonyms
  or synonyms of each other) into compositional networks.
\item Our method addresses generalization tasks in gSCAN which no prior work does,
     such as learning from a few examples and generalizing to longer sequences.
  
  

  
  
\end{compactenum}


\section{Related Work}

\paragraph{Command following}
Robots must ground in their surroundings.
Previous work grounds concepts such as objects~\cite{guadarrama2014open},
spatial relations, and object properties~\cite{kollar2010toward}.
To turn a command into actions, \citet{chen2011learning} and \citet{matuszek2013Learning}
learn semantic parsers that convert instructions into plans.
\citet{mei2016listen} demonstrate a seq2seq network fused with a visual encoder to
predict action sequences from input sentences.
This type of seq2seq network is adopted by many supervised models and reinforcement
learning agents~\cite{fu2019language, shah2018follownet}.
\citet{blukis2018Mapping} present a U-Net architecture that predicts goal distributions 
conditioned on linguistic commands to control a drone.
Predicting a single final goal may not always be ideal as language can describe
the manner of interacting with objects \& the world.
\citet{kuo2020language} demonstrate that a compositional network structured according to the
parse of the input command can combine with a sampling-based motion planner to guide the
sampling process.
Similar to Kuo \etal, we use RNNs as the base units of the model and compose networks
from parses.
Our approach is further compatible with any type of parsers and can encode lexical 
semantics of words, which allows us to investigate how compositional architectures
generalize systematically.

\paragraph{Generalization in grounded language understanding}
Many methods have been proposed to test an agent's generalization capabilities in
different perspectives of grounded language understanding.
\citet{yu2018interactive} consider a multi-task setting and train an agent to navigate
a 2D maze and to answer grounded questions.
\citet{pezzelle2019MALeViC} focus on evaluating agents' abilities in assessing the
meaning of adjectives in context.
%
%
\citet{chaplot2018gated} and \citet{hermann2017grounded} evaluate RL agents' capability to
generalize to novel composition of shape, size, and color in 3D simulators.
The BabyAI platform~\cite{chevalier2018babyai} evaluates RL agents in a grid world with
tasks that demand an increasing understanding of the compositional structure of their domain.
They show that RL agents generalize poorly when the tasks have a compositional structure.
\citet{Bogin2021Composition} learn latent trees to ground compositional reasoning in
the visual question answering domain.
Rather than focusing on one aspect of generalization as much of the prior work does,
gSCAN~\cite{ruis2020gscan} takes ideas from meaning composition to create a systematic
battery of tests for generalizing in grounded settings.
%
%
A few recent approaches attempted to address the generalization challenges in gSCAN.
\citet{heinze2020think} add an auxiliary loss in the baseline seq2seq model to predict the 
location of the target object.
However, it only improves in a few subsets related to target object predictions.
\citet{gao2020systematic} use a language conditioned graph network to model the relation 
between the objects and natural-language context.
While the graph network improves some subsets of novel compositions, they did not
evaluate on few-shot learning and generalization to longer action sequences.

\paragraph{Compositional networks}
The idea that linguistic structures and compositionality can be reflected in the
internal workings of a model to enable better generalization is not itself new
\citep{liang2015bringing}.
\citet{tellex2011understanding} and \citet{barbu2012video} mirror the linguistic
structures produced by a constituency parser in the structure of a graphical
model to respectively execute robotic commands and recognize actions.
Similarly, \citet{socher2011parsing} and \citet{legrand2014joint} build neural networks based on parse trees.
\citet{andreas2016learning} demonstrate a procedure to compose a collection of
network modules based on a semantic parser for visual question answering.
Not all modular networks are derived from language; for example, prior work has
modularized sub-policies and sub-goals in embodied question
answering~\cite{eqa_modular} or transfer learning~\cite{alet2018modular}
according to other task-specific principles.
%
%


\section{Technical Approach}
\label{sec:approach}

We first describe how the compositional networks can be constructed from any
linguistic parses.
Then, we show how a linguistic notion, such as a known relationship between
words, can be incorporated in the model.

\subsection{Parsing natural-language commands}

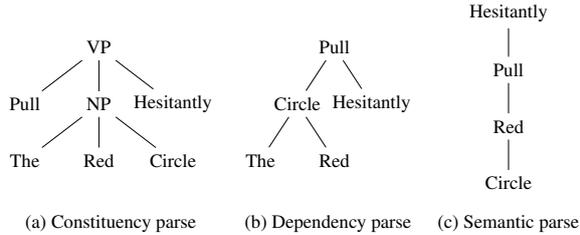
\begin{figure}
    \centering
    \scalebox{0.65}{
    \begin{tabular}{ccc}
      \begin{tikzpicture}[level distance = 3em]
        \node[white] (start) {};
        \node[black, above=2.5 of start] (vp) {VP} {
          child {node [black] (pull) {Pull}}
          child {node [black] (np) {NP}
            child {node [black] (the) {The}}
            child {node [black] (red) {Red}}
            child {node [black] (circle) {Circle}}
          }
          child {node [black] (hes) {Hesitantly}}
        };
      \end{tikzpicture} &
      \begin{tikzpicture}[level distance = 3em]
        \node[white] (start) {};
        \node[black, above=2.5 of start] (pull) {Pull} {
          child {node [black] (circle) {Circle}
            child {node [black] (the) {The}}
            child {node [black] (red) {Red}}
          }
          child {node [black] (hes) {Hesitantly}}
        };
      \end{tikzpicture} &
      \begin{tikzpicture}[level distance = 3em]
        \node[black] (hes) {Hesitantly} {
          child {node [black] (pull) {Pull}
            child {node [black] (red) {Red}
              child {node [black] (circle) {Circle}}
            }
          }
        };
      \end{tikzpicture}\\[1ex]
      (a) Constituency parse & (b) Dependency parse & (c) Semantic parse\\
    \end{tabular}
    }
    \caption{\small{Parses for the command ``Pull the red circle hesitantly.'' in three
      formalisms. Each leads to different compositional networks which have
      radically different generalization abilities.}}
    \label{fig:parsers}
    \vspace{-2ex}
\end{figure}

Given a natural-language command, a parser produces a hierarchical structure
of that command revealing its part-based compositional structure, \ie which words
modify one another, and the nature of that modification.
Different approaches to analyzing linguistic utterances lead to different structures.
Here we consider three kinds of parsers: a constituency parser \citep{joshi2018extending},
a dependency parser \citep{dozat2016deep}, and a semantic parser; see
Figure~\ref{fig:parsers} for an example of the different structures produced.
In what follows, we use the language of constituency parsing: that a parse is a
collection of nodes arranged in a tree; dependency parses consist of words and
binary relationships between words; while semantic parses in this work consist of
a formula in propositional logic.
This is purely for linguistic convenience, as no shared lexicon exists between
these parsers.
Our approach treats all parses as labeled directed acyclic graphs and is
agnostic to the source of the parse.
In the Results, we discuss the differences between these parsers.

\subsection{Constructing compositional networks}

Given the parse of a command, the nodes in the parse tree are replaced with RNNs 
connected to one another according to the structure of the parse.
An example network structure is shown in Figure~\ref{fig:model}.
This compositional network is used to predict actions for the agent to follow
based on the visual observation at every time step.

\paragraph{Recurrent word modules}
We use RNNs as the basic building blocks of the compositional networks because
the hidden states provide the capacity to maintain the context of task
progression, for example, pushing heavy objects twice in order to move them.
Each word or predicate/function in the semantic parse corresponds to a specific
RNN, forming a lexicon of RNNs.
In the case of dependency or constituency parses, we create a separate model for
each word depending on the arity of that word in the parse tree.
Most parses are trees as described above, rooted in one node, corresponding to
one word, predicate, or operator in the parse.
Some parses can consist of multitrees, one or more trees that can share nodes.
In this case, we can synthesize a dummy root node.
Note that this operation of inserting a dummy root has linguistic precedent; for
example, dependencies are considered by some to have a phantom root
\citep{ballesteros2013going}.
The word that is the root plays a special role: its hidden state is decoded by
a linear layer that computes a distribution over the next action.

\paragraph{Connecting word modules}
The information that flows between nodes always follows the reverse direction of the
arcs in the parses.
In the cases of parse trees described above, the information flows from children to
its parent, \ie from leaves to the root.
Labels on the arcs are used to keep consistent the input to nodes with more than one
argument.
For example, the word ``grab'' usually involves two arguments, the agent and the
patient; the RNN for ``grab'' takes as input the output of the RNN that corresponds
to the agent first and the one for the patient second, consistently.
Words with multiple arguments use a linear layer to combine together the input
embeddings; the arc labels determine the arbitrary but consistent order in which
the multiple input vectors should be combined before this linear layer.

\paragraph{Attention mechanism in word modules}
Within each word module, the RNN maintains a state vector and this internal
representation is always used to predict an attention map before being accessed by other
word modules.
At each time step $t$, the module for word $w$ receives as input an embedding 
$\textit{obs}_t$ of the agent's surroundings, the attention maps from its children 
$\textit{att}_t^{c_1} \cdots \textit{att}_t^{c_n}$, and its own state vector $h_{t-1}^w$
from the previous time step.
The embedding $\textit{obs}_t$ is computed by a CNN which is co-trained with the rest
of the network.
The attention map for word $w$ is computed as follows:
\begin{align*}
    \textit{att}_t^w = \textit{softmax}(\textit{MLP}(\textit{h}_{t-1}^w,
        \textit{obs}_t \odot \textit{att}_t^{c_1}, \\
        \cdots, \textit{obs}_t \odot \textit{att}_t^{c_n}))
\end{align*}
The observation is weighted by the attention maps from children first and combined
with the hidden state to predict where to attend, \ie the meaning of a word is
grounded in the map.
Inside the $\textit{MLP}$, the weighted observations and the hidden state are
mapped to the same dimension before being combined together.
The attention map is normalized with softmax and adds up to 1.
The RNN then takes this attention map to update its hidden state:
\begin{equation*}
    o_t^w, h_t^w = \textit{RNN}^w(\textit{obs}_t \odot \textit{att}_t^{w}, h_{t-1}^w)
\end{equation*}
Attention maps are the only mechanism by which nodes communicate with one
another.
We demonstrate in the Results that this is critical to performance.
It provides a common representation for all words, which in a sense makes all
words compatible with one another.
Without this restriction, words might never develop the ability to understand
one another's representations.

\paragraph{Training compositional networks}
We train the CNN to encode the observations, the RNNs and attention modules for each word jointly.
At training time, the input consists of pairs of commands and corresponding
trajectories.
The parser is pretrained, and in the case of the constituency and dependency
parsing, an off-the-shelf general-purpose English model is used.
The command is parsed and a corresponding network is instantiated.
The word modules that have not been discovered in previous commands are instantiated
with random weights.
No information is provided as to which which part of the trajectory and
relationships between the trajectory and other objects, and what each word in the
command might refer to.
The parameters of the resulting compositional network are trained without knowing
the mapping between words and meanings.
This knowledge must be inferred during training, thereby disentangling the meanings
of each word.
During training, the agent is provided with the ground-truth action at each time
step to compute the maximum log-likelihood loss of the distribution over the
next action and update the network.

\subsection{Incorporating lexical semantics}
Humans bring to bear tremendous prior knowledge to any new learning problem.
\citet{dubey2018investigating} show that depriving humans of that knowledge by,
for example, making dangerous objects look safe and vice versa,
significantly impairs the ability of humans to learn and generalize.
To this end, we demonstrate how to naturally add weak constraints, automatically
derived from WordNet~\citep{fellbaum2012wordnet}, about the meanings of words.

Given a token in a parse, we search WordNet for related synonyms and antonyms.
When creating the lexicon of RNNs, we consider the transitive closure of
synonyms and antonyms as a single RNN for that concept.
The combined RNN, \eg ``Small\&Large'' RNN in Figure~\ref{fig:model}, has two
attention map outputs, but only one of the two is used depending on which 
variant of the concept appeared in the input.
Intuitively, the computations to determine the relative sizes of objects are
closely related to one another, regardless of whether one is checking if an
object is small or large; this approach shares those computations between
synonyms and antonyms.
Critically, at training time, we add an additional loss, that the attention maps
of these two concepts should be inverse of one another.
This is done by optimizing the negative Hausdorff distance, which for grayscale
maps minimizes the total intensity in the product of the two attention maps.
A simple negation of maps would be ineffective as it would force one concept to be true
when the other is not, which is not what being an antonym means.
Not all objects that are not small, are large; some are merely irrelevant or their
size is indeterminate.
But, relative to a single reference object, the same object cannot usually be
large and small at the same time.
Hence, during training time, we add an auxiliary loss by computing the negative Hausdorff distance of attention maps of antonyms.
This loss is used to avoid both attention maps
paying attention to the same regions without disturbing one another when one of
the two concepts is irrelevant.
In general, knowledge about relationship between words can be used to
augment the network, perhaps as derived from word embeddings.

\section{Experiments}
\label{sec:result}

We evaluate the compositional network on the gSCAN dataset~\cite{ruis2020gscan}
which was designed to systematically test the generalization ability of grounded
agents.
Our vocabulary size and trajectory distributions are the same as in gSCAN.
The observation space for the agent is a $6 \times 6$ grid and the agent can choose from six random actions: walk, turn left/right, push, pull, and stay.

\begin{figure}
  \centering
  \scalebox{0.75}{
  \begin{tabular}{cc}
    \includegraphics[width=0.25\textwidth]{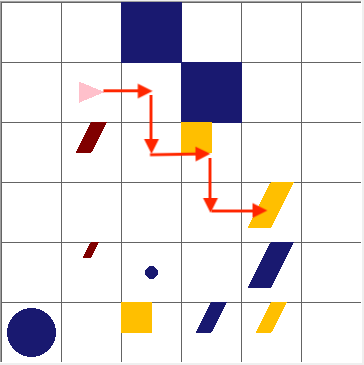}
    &\includegraphics[width=0.25\textwidth]{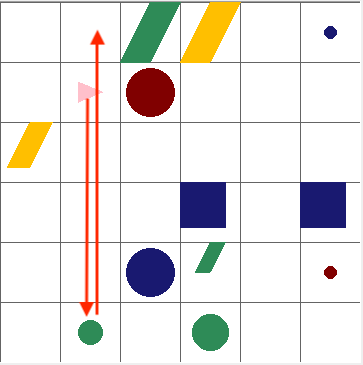}\\
    (a) Walk to a big yellow cylinder & \\[-0.8ex]
    while zigzagging & \raisebox{1.3ex}{(b) Pull a small green circle}
  \end{tabular}
  }
  \caption{\small{Two examples from gSCAN. The pink triangle is the agent with
    the tip of the triangle pointing forward. Red arrows show a trajectory.
    (a) A sentence that contains an action modifier. When testing novel adverb-verb
    combinations, the agent might
    separately see the concept ``walk'' and the concept ``while zigzagging'' in
    different sentences, but must infer what to do when concepts are
    combined during testing. (b) The agent must
    understand the target object, but size is relative. What is large on one
    map, might be small on another, depending on what other objects are
    available. In these test conditions, certain object sizes never appear
    labeled as large or small; this must be inferred from the context and then
    generalized to new sizes.}}
  \label{fig:gscan-example}
  \vspace{-3ex}
\end{figure}

\begin{table*}
  \centering
  \scalebox{0.65}{%
  \begin{tabular}{@{}l@{\hspace{-0.5ex}}c@{}c@{}c@{}c@{}c@{}c@{}c@{}c@{}c@{}}
        & \multicolumn{1}{c}{\parbox[t]{2.4cm}{Seq2seq}}
        & \multicolumn{1}{c}{\parbox[t]{2.4cm}{GECA}}
        & \multicolumn{1}{c}{\parbox[t]{2.4cm}{AuxLoss}}
        & \multicolumn{1}{c}{\parbox[t]{2.4cm}{LCGN}}
        & \multicolumn{1}{c}{\parbox[t]{2.0cm}{State\\Ours}} &
    \multicolumn{1}{c}{\parbox[t]{2.0cm}{+Attention\\Ours}} &
    \multicolumn{1}{c}{\parbox[t]{2.0cm}{Constituency\\Ours}} &
    \multicolumn{1}{c}{\parbox[t]{2.0cm}{Dependency\\Ours}} &
    \multicolumn{1}{c}{\parbox[t]{2.0cm}{Semantic\\Ours}}\\
    \cmidrule(lr){2-2} \cmidrule(lr){3-3} \cmidrule(lr){4-4} \cmidrule(lr){5-5} \cmidrule(lr){6-6} \cmidrule(lr){7-7} \cmidrule(lr){8-8} \cmidrule(lr){9-9} \cmidrule(lr){10-10}
    A   & 97.69 $\pm$ \phantom{0}0.22 & 87.60 $\pm$ \phantom{0}1.19 & 94.19 $\pm$ \phantom{0}0.71 & \textbf{98.60} $\pm$ \phantom{0}0.95 & 49.83 $\pm$ \phantom{0}5.05 & 96.06 $\pm$ \phantom{0}1.40 & 96.20 $\pm$ \phantom{0}1.68 & 96.91 $\pm$ \phantom{0}1.86 & 96.73 $\pm$ \phantom{0}0.58 \\
    B  & 54.96 $\pm$ 39.39 & 34.92 $\pm$ 39.30 & 86.45 $\pm$ \phantom{0}6.28 & \textbf{99.08} $\pm$ \phantom{0}0.69 & \phantom{0}4.37 $\pm$ \phantom{0}2.90 & 79.36 $\pm$ 32.71 & 80.82 $\pm$ \phantom{0}7.34 & 58.42 $\pm$ 18.31 & 94.91 $\pm$ \phantom{0}1.30 \\
    C     & 23.51 $\pm$ 21.82 & 78.77 $\pm$ \phantom{0}6.63 & 81.07 $\pm$ 10.12 & \textbf{80.31} $\pm$ 24.51 & \phantom{0}5.53 $\pm$ \phantom{0}1.75 & 43.93 $\pm$ 15.42 & 40.33 $\pm$ \phantom{0}7.63 & 64.23 $\pm$ \phantom{0}6.04 & 67.72 $\pm$ 10.83 \\
    D & \phantom{0}0.00 $\pm$ \phantom{0}0.00 & \phantom{0}0.00 $\pm$ \phantom{0}0.00 & - & \phantom{0}0.16 $\pm$ \phantom{0}0.12 & \phantom{0}1.62 $\pm$ \phantom{0}0.79 & \phantom{0}3.41 $\pm$ \phantom{0}1.21 & \phantom{0}3.66 $\pm$ \phantom{0}2.93 & \phantom{0}5.29 $\pm$ \phantom{0}3.36 & \textbf{11.52} $\pm$ \phantom{0}8.18 \\
    E      & 35.02 $\pm$ \phantom{0}2.35 & 33.19 $\pm$ \phantom{0}3.69 & 43.43 $\pm$  \phantom{0}7.0 & \textbf{87.32} $\pm$ 27.38 & 27.18 $\pm$ \phantom{0}3.75 & 68.84 $\pm$ 34.72 & 52.96 $\pm$ 15.19 & 28.34 $\pm$ 16.13 & 76.83 $\pm$ \phantom{0}2.32 \\
    F & 92.52 $\pm$ \phantom{0}6.75 & 85.99 $\pm$ \phantom{0}0.85 & - & \textbf{99.33} $\pm$ \phantom{0}0.46 & 43.29 $\pm$ \phantom{0}1.90 & 90.09 $\pm$ 14.81 & 97.25 $\pm$ \phantom{0}0.17 & 96.99 $\pm$ \phantom{0}1.79 & 98.67 $\pm$ \phantom{0}0.05 \\
    G k=1                 & \phantom{0}0.00 $\pm$ \phantom{0}0.00 & \phantom{0}0.00 $\pm$ \phantom{0}0.00 & - & - & \phantom{0}3.38 $\pm$ \phantom{0}3.66 & \textbf{\phantom{0}1.79} $\pm$ \phantom{0}0.69 & - & - & \phantom{0}1.14 $\pm$ \phantom{0}0.30 \\
    \phantom{G }k=5       & \phantom{0}0.47 $\pm$ \phantom{0}0.14 & - & - & - & \phantom{0}4.87 $\pm$ \phantom{0}1.22 & \phantom{0}6.31 $\pm$ \phantom{0}5.66 & - & - & \textbf{\phantom{0}8.85} $\pm$ \phantom{0}1.87 \\
    \phantom{G }k=10      & \phantom{0}2.04 $\pm$ \phantom{0}0.95 & - & - & - &  \phantom{0}8.48 $\pm$ \phantom{0}4.72 & 34.28 $\pm$ \phantom{0}6.59 & - & - & \textbf{36.91} $\pm$ \phantom{0}5.13 \\
    \phantom{G }k=50      & \phantom{0}4.63 $\pm$ \phantom{0}2.08 & - & - & - & 13.19 $\pm$ \phantom{0}2.53  & 45.79 $\pm$ 13.53 & - & - & \textbf{46.30} $\pm$ 11.69 \\
    H  & 22.70 $\pm$ \phantom{0}4.59 & 11.83 $\pm$ \phantom{0}0.31 & - & \textbf{33.60} $\pm$ 20.81 & \phantom{0}9.80 $\pm$ \phantom{0}0.74 & 13.27 $\pm$ \phantom{0}8.75 & 20.84 $\pm$ \phantom{0}1.87 & \phantom{0}0.00 $\pm$ \phantom{0}0.00 & 20.98 $\pm$ \phantom{0}1.38 \\
    I & \multicolumn{8}{l}{See table \ref{res:longseq}; only the original publication and this work address generalization condition I.}
  \end{tabular}
}
  \caption{\small{Performance on gSCAN including models from the original publication Seq2seq and
  GECA \citep{ruis2020gscan,andreas2019good} as well as other recent models
  AuxLoss \citep{heinze2020think} and LCGN, the language conditioned graph
  network \citep{gao2020systematic}.  The first row, condition A, does not represent generalization
  performance; it is the performance when the training and testing sentence distributions
  are the same. AuxLoss, LCGN, and our work are able to generalize to B and C.
  Our model is the only one to show any generalization in D.
  LCGN and our model have similar performance on E; note the very high variance of LCGN.
  Our model is the only one that addresses generalization condition G aside from Seq2Seq.
  While LCGN outperforms our model in H, we note its extremely high variance.
  No other work addresses generalization condition I.}}
  \label{res:composition}
  \vspace{-2ex}
\end{table*}

Figure~\ref{fig:gscan-example} shows examples of two gSCAN commands in different
environments.
At test time, an agent receives a command and an environment
(randomly placed objects with random sizes and colors).
It predicts a sequence of actions to carry out that command.
gSCAN includes adjectives that describe an object's color and size, nouns, verbs,
prepositional phrases, and adverbs.
We summarize the generalization conditions in gSCAN below.
\begin{compactenum}[A]
\item \emph{Random}: all concepts and combinations appear in the training set
  to put other results in context.
\item \emph{Yellow squares} holds out types of references to an object,
  \eg it is never referred to as ``yellow square'' but only as ``small square''.
\item \emph{Red squares} holds out any references to an object, \eg red squares are
  never referenced.
\item \emph{Novel direction} never refers to a object in a selected direction, 
  \eg the target is located at south-west of the agent.
\item \emph{Relativity} never refers to objects with a given relative size, \eg what is small while training may be large when testing.
\item \emph{Class inference} requires inferring unstated properties,
  \eg object size determines how many \textsc{pull} actions are required to move it.
\item \emph{Adverbs} requires learning a word such as ``cautiously'' from a small
  given number of examples.
\item \emph{Adverb to verb} holds out pairs of verbs and
  action modifier, \eg ``walking'' while ``spinning''.
\item \emph{Sequence length} generalizes to longer action sequences.
\end{compactenum}

\subsection{Models}

We evaluate several variations of our compositional networks
\footnote{Source code is available at \\https://github.com/ylkuo/compositional-gscan}
against baseline models described in \citet{ruis2020gscan} (a seq2seq model and GECA
introduced in \citet{andreas2019good}) as well as two recent models discussed in the
Related Work \citep{heinze2020think,gao2020systematic}.
The seq2seq model encodes both the commands and the environment separately using a
BiLSTM and a CNN.
This is a common architecture used in many publications.
GECA is a variant of the baseline seq2seq model which employs data augmentation
to improve generalization.

We consider three variants of our full model, each using different parsers to
structure the compositional networks.
We use a pretrained constituency parser from AllenNLP~\cite{Gardner2017AllenNLP};
a pretrained dependency parser from Stanza~\cite{qi2020stanza}; and a semantic
parser which rewrites the original grammar used to create gSCAN.
These three models communicate using attention maps.
All compositional networks presented in the evaluation contain a CNN with
kernel size 7 and 50 channels and are trained with lexical semantics.
Each word module is a GRU with 2 hidden layers and 20-dimensional hidden states.
A component uses a linear layer to map the input observation and hidden state to
dimension of 10, and the ReLU activation in MLP to predict the grayscale attention for each 
grid cell.
We select hyperparameters that increase exact matches in the validation set.
We train all networks using the Adam optimizer with the initial learning rate 0.001,
$\beta_1$ 0.9, and $\beta_2$ 0.999.

To demonstrate how critical a mechanism that makes modules mutually
intelligible to one another when testing compositionality is, we test two other
models.
The first is a model that passes a 20-dimensional state vector instead of an attention map;
it is referred to as \emph{State}.
Since the capacity of an attention map and an $n$-dimensional state vector
cannot be matched, no matter what value $n$ takes, we give the next variant an
even more powerful representation.
\emph{+Attention} includes both the state vector and the attention map.
This is strictly more powerful but fails to generalize well because other models
cannot understand this side channel muddling the information being exchanged.
Both ablations employ the semantic parser to compose the networks.

\begin{table*}
  \centering
  \scalebox{0.7}{%
  \begin{tabular}{ccccc}
       & Seq2seq~\citep{ruis2020gscan} & \multicolumn{3}{c}{Ours w/ Semantic parses} \\
    Target length & \multicolumn{1}{c}{Train length $\leq 15$} &  \multicolumn{1}{c}{Train length $\leq 15$} & \multicolumn{1}{c}{Train length $\leq 14$} & \multicolumn{1}{c}{Train length $\leq 13$}\\
    \cmidrule(lr){2-2} \cmidrule(lr){3-5}
    \
    $\phantom{\geq }15$ & 94.98 $\pm$ 0.12 & 92.06 $\pm$ 1.94 & 75.02 $\pm$ 8.33 & 66.11 $\pm$ 8.46 \\
    $\phantom{\geq }16$ & 19.32 $\pm$ 0.02 & 89.05 $\pm$ 2.60 & 67.39 $\pm$ 9.61 & 59.93 $\pm$ 9.75 \\
    $\phantom{\geq }17$ & \phantom{0}1.71 $\pm$ 0.38 & 85.08 $\pm$ 4.02 & 62.43 $\pm$ 9.84 & 56.14 $\pm$ 11.06 \\
    $\geq 18$           & $\phantom{0}< 1$  & 53.67 $\pm$ 4.00 & 34.01 $\pm$ 9.78 & 31.33 $\pm$ 10.28 \\
  \end{tabular}
  }
  \caption{\small{Performance on gSCAN as a function of the length of action sequences in the
    training set. State-of-the-art methods fail to generalize to longer sequences. Our model does, although not perfectly. Even as the training action sequence length is decreased, our model continues to generalize.}}
  \label{res:longseq}
  \vspace{-3ex}
\end{table*}

\subsection{Results}

Experiments were carried out on two machines, each with  80-core Intel Xeon 6248 2.5GHz CPUs,
768 GB of RAM, and 8 Titan RTX 24GB GPUs.
Training and testing the models including ablations took approximately four days.
Each model trained for 150,000 steps with batch size 200.
Table~\ref{res:composition} summarizes percentage of exact match and the
standard deviation over three runs by generalization condition.
Overall, the model using the semantic parser and attention maps significantly
outperforms all other variants.
An arbitrary state vector, \emph{State}, passed between words performs very
poorly, far worse than the non-compositional seq2seq model.
It appears to be critical that there exists a method to make representations
interpretable to models which have not been exposed to one another.
Adding attention maps to the state, \emph{+Attention}, results in better
performance but still worse than passing attention maps only.

Only when we remove all arbitrary state and only exchange attention maps does
compositionality shine through.
The three models in the rightmost three columns of Table~\ref{res:composition}
generalize in most conditions.
In cases where prior work such as AuxLoss and LCGN demonstrate generalization,
our models achieve state of the art or close to state of the art performance.
In cases such as conditions, D, G, and, as will be shown later, I, our model
generalizes when others do not.
AuxLoss and LCGN do not report results on G and I.
Note that, our results in condition G show that our model only needs a handful of examples to achieve reasonable performance.
This capability allows our model to scale to novel words and objects more quickly.

%

In most cases, networks based on the semantic parses outperform those based
on syntactic parses.
This may be because semantic parses are more stable than syntactic parses,
\ie similar concepts can have very different surface representations but their
relationship is revealed in a deeper analysis.
It could be that this phenomenon occurs for a much more interesting reason:
semantic parses are designed to be useful for extracting the meaning of
sentences.
Perhaps, in the future, grounded agents can provide a completely independent
and novel test for linguistic representations --
a good representation is one where a robot is able to learn to perform well.

gSCAN includes a condition, I, that extends the dataset to longer sequences.
Table~\ref{res:longseq} summarizes our performance on this condition comparing with Ruis \etal~\cite{ruis2020gscan}.
Note that other models do not report results for condition I.
%
We have retrained Gao \etal's model ~\cite{gao2020systematic} for condition I and received $1.36 \pm 0.34$ exact match on the test set for over three runs. 
When the training sequence length and the test sequence length are the same, our
model performs well, in line with the state of the art.
As the sequence length increases, baseline seq2seq models lose all of their performance
almost immediately.
The performance of our compositional model does decay, but at a far slower rate.
We can train with even shorter sequence lengths, 13 instead of 15, and still vastly
outperform the state of the art at predicting move sequences of length 18.

\subsubsection{Interpretability and acquisition}

\begin{figure*}[t]
  \centering
  \vspace{3ex}
  \scalebox{0.9}{%
  \begin{small}
    \begin{tabular}{ccccccc}
      \includegraphics[width=0.12\linewidth,frame]{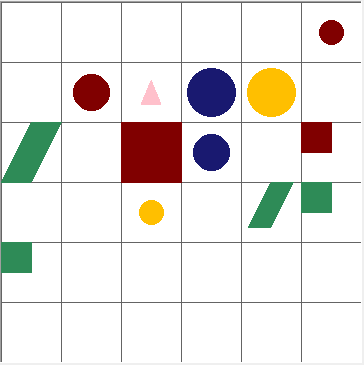}
      &\includegraphics[width=0.12\linewidth,frame]{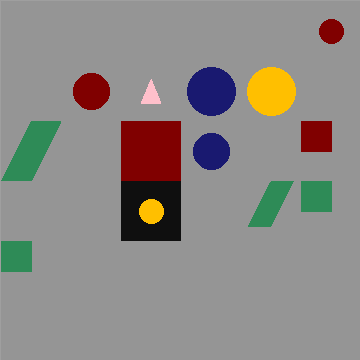}
      &\includegraphics[width=0.12\linewidth,frame]{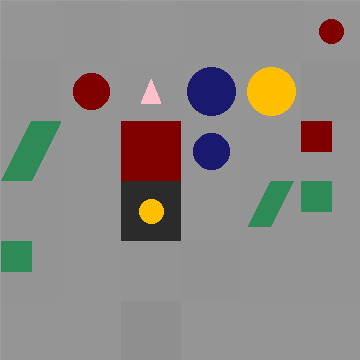}
      &\includegraphics[width=0.12\linewidth,frame]{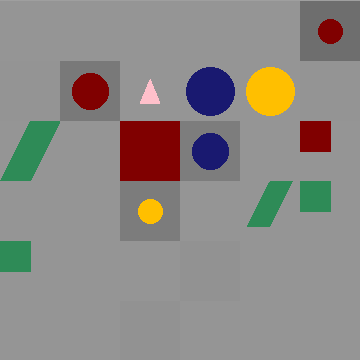}
      &\includegraphics[width=0.12\linewidth]{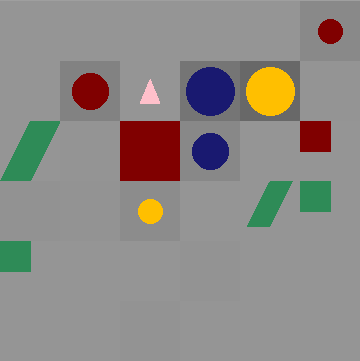}
      &\includegraphics[width=0.12\linewidth]{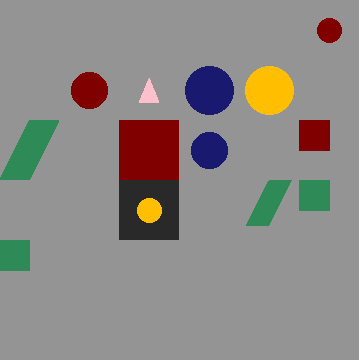}
      &\includegraphics[width=0.12\linewidth]{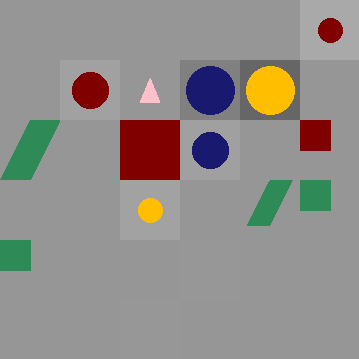}\\
      \emph{Initial environment} & \emph{Walk} & \emph{Yellow} & \emph{Small} & \emph{Circle} & \emph{While spinning} & \emph{Size ordering} \\
    \end{tabular}
  \end{small}
  }
  \vspace*{-1.0ex}
  \caption{\small{Attention maps while executing \emph{Walk to a small yellow circle while spinning}.
  Darker cells are areas of interest to models. Models refine the attention maps they receive as input
  from their children, \eg \emph{circle} attends to all circles, ``small'' filters there to small
  circles, and ``yellow`` focuses on the combination of all three. Since the models are informed by
  lexcial semantics (small and big are antonyms), we can infer the size ordering map, where lightness
  correlates with circle size.}}
  \label{fig:attention}
  \vspace*{-1.5ex}
\end{figure*}

\begin{figure*}
  \centering
  \scalebox{0.8}{%
  \begin{small}
    \begin{tabular}{cccccc}
      & $t=0$ & $t=4$ & $t=8$ & $t=12$ & $t=16$ \\
      \raisebox{6ex}{\emph{Walk}} & \includegraphics[width=0.12\linewidth,frame]{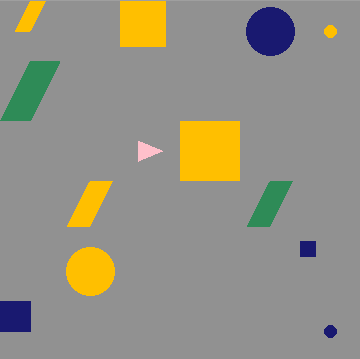}
      &\includegraphics[width=0.12\linewidth,frame]{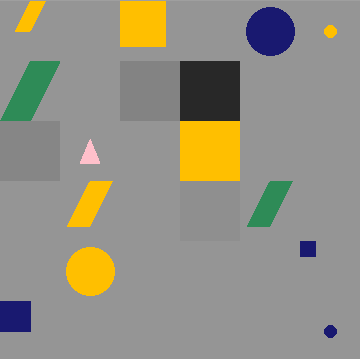}
      &\includegraphics[width=0.12\linewidth,frame]{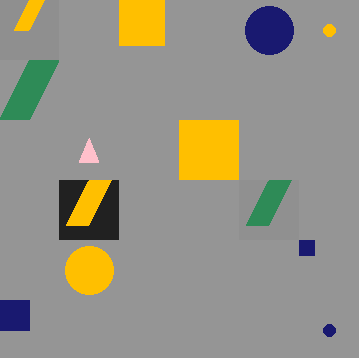}
      &\includegraphics[width=0.12\linewidth]{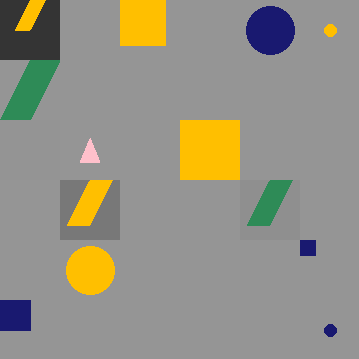}
      &\includegraphics[width=0.12\linewidth]{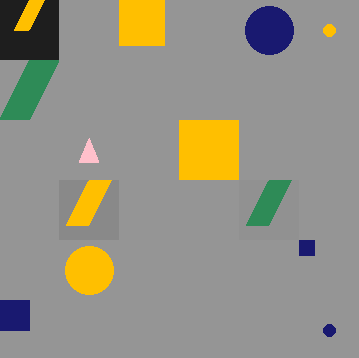} \\
      \raisebox{6ex}{\emph{Small}} & \includegraphics[width=0.12\linewidth,frame]{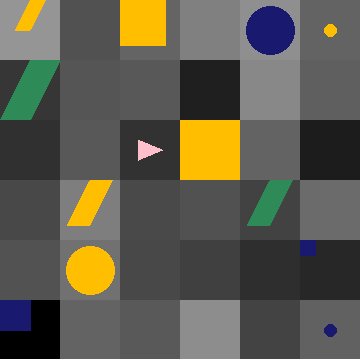}
      &\includegraphics[width=0.12\linewidth,frame]{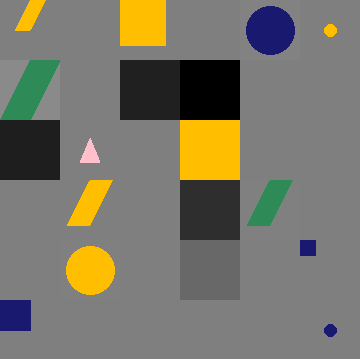}
      &\includegraphics[width=0.12\linewidth,frame]{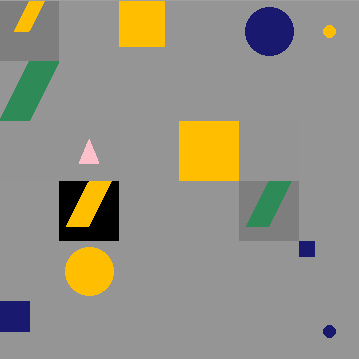}
      &\includegraphics[width=0.12\linewidth]{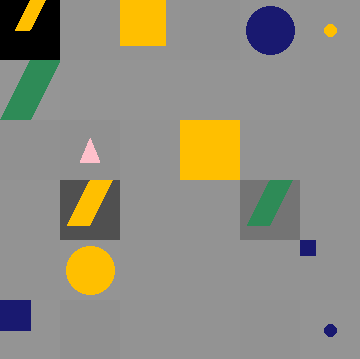}
      &\includegraphics[width=0.12\linewidth]{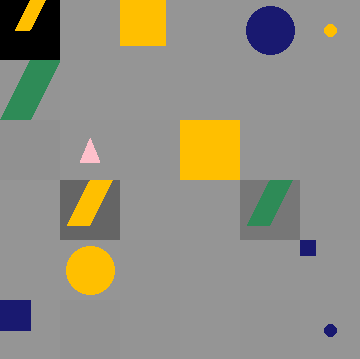} \\
      \raisebox{6ex}{\emph{Cylinder}} & \includegraphics[width=0.12\linewidth,frame]{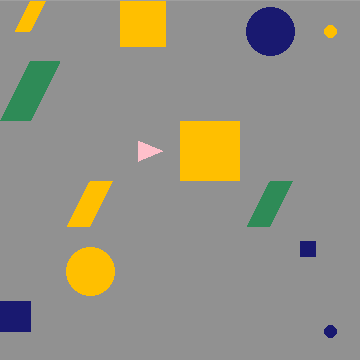}
      &\includegraphics[width=0.12\linewidth,frame]{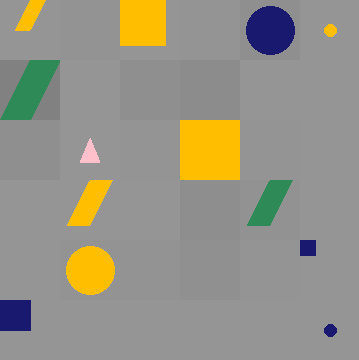}
      &\includegraphics[width=0.12\linewidth,frame]{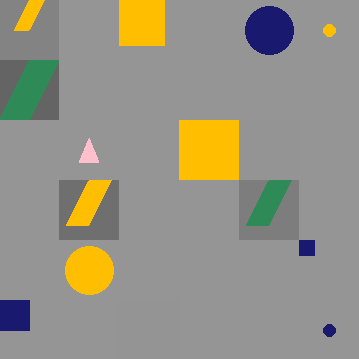}
      &\includegraphics[width=0.12\linewidth]{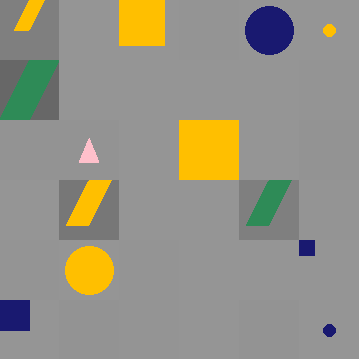}
      &\includegraphics[width=0.12\linewidth]{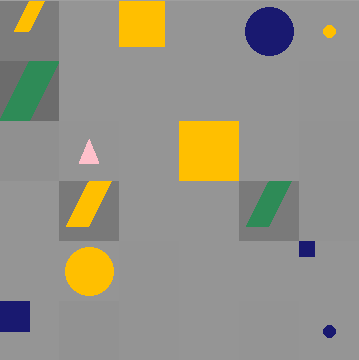} \\
      \raisebox{6ex}{\emph{While spinning}} & \includegraphics[width=0.12\linewidth,frame]{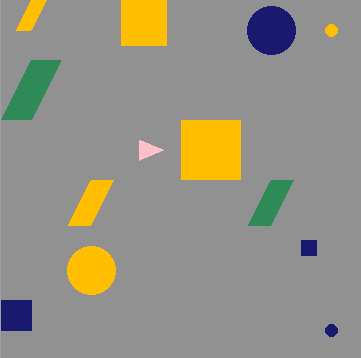}
      &\includegraphics[width=0.12\linewidth,frame]{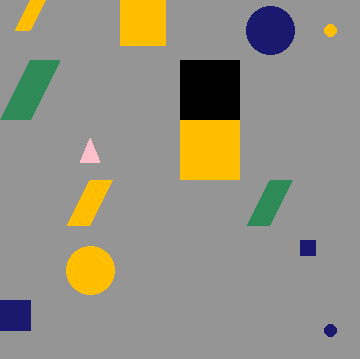}
      &\includegraphics[width=0.12\linewidth,frame]{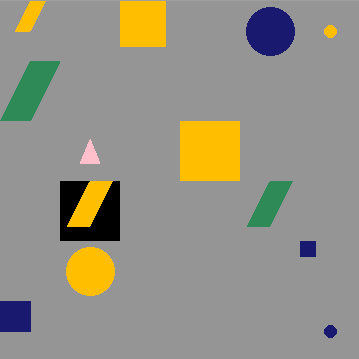}
      &\includegraphics[width=0.12\linewidth]{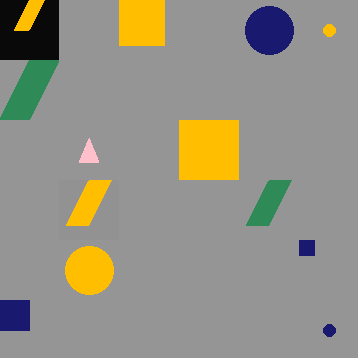}
      &\includegraphics[width=0.12\linewidth]{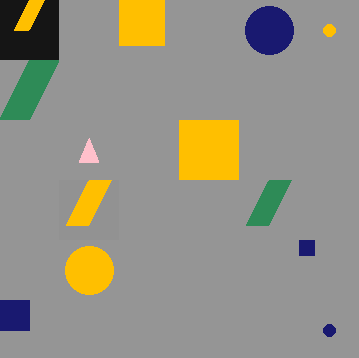} \\
    \end{tabular}
  \end{small}
  }
  \vspace*{-1.5ex}
  \caption{\small{Visualization of the learning progression for the command ``Walk to a small
  cylinder while spinning.'' Each row contains the attention maps produced by the word modules
  at different training epochs. The maps on the left are the beginning of the
  training, where the attentions are uniform or random. Toward the right, as the training
  epochs increase, the ``Cylinder'' module first identifies the shapes of the objects, and
  then the ``Small'' module takes longer time to learn to sort by the size of cylinders.
  By inspecting the attention maps over time, we can track if a module acquires the meaning
  of the word and what it is confused about, for example, the ``Small'' module at $t=8$ can
  identify the smaller cylinders but confused about the size ordering.}}
  \label{fig:learning}
  \vspace*{-2.5ex}
\end{figure*}

Our model is interpretable in two ways.
(1) The structure of the network overtly encodes the structure of the sentence so a
parser error can be observed directly.
(2) The internal reasoning of the network proceeds by passing attention maps between
modules. 
These maps can be directly inspected to see what different words or phrases are
physically referring to.
If an agent picks up the wrong object because the words that refer to that object 
attend to the wrong part of the map, this error will be evident from the attention maps.
Figure~\ref{fig:attention} shows example attention maps which can be viewed as a series
of selectors to filter the goal object to interact with.
Furthermore, since the Small\&Large attention maps are cotrained to have the opposite
semantics, we can use them to infer the absolute scale of object sizes by post-processing
the two maps: $(-\textit{att}_\textit{Small}+\textit{att}_\textit{Large})/2$.
We can also inspect the attention maps across training epochs to see if the network
acquires the meaning of the word and how the representations change over time.
Figure~\ref{fig:learning} demonstrates the learning progression of the network through
attention maps.

\section{Conclusion}
\label{sec:conclusion}

We have presented a model that addresses many of the compositionality
challenges found in gSCAN, a dataset designed to challenge networks.
%
%
When the compositionality inherent in a problem is reflected in the computation
of a network, the resulting network is far better able to understand the target
domain.
This is only critical at test time, when generalizing to new combinations.
An important caveat is that a mechanism for making representations of
different word modules compatible with one another is key.
Here we do this by constraining all communication through attention maps.

Performance of compositional approaches depends on what is being composed
and how.
When the compositionality does not capture part of a problem, such as
condition D here, it does not meaningfully improve results.
When compositionality is relevant, it appears that it can supplant data
augmentation and provide a faster, principled, dataset-agnostic method
to achieve better results.
When compositionality is derived from language, it enables the inclusion of
linguistic notions, \eg synonyms and antonyms, in models.

The most suggestive and admittedly tenuous implication of this work is that
perhaps we can use this approach to test linguistic representations.
Many formalisms exist in linguistics for encoding semantics.
Without an independent test for which is better, convergence to one formalism is unlikely.
It appears that when compositional models are trained to perform tasks, some
representations are significantly better than others.
In our experiment, abstract representations, \ie ones further from the surface syntax of
language, result in better models.
Perhaps in the future a meta-learning approach could allow grounded robotics to come
full circle: from borrowing ideas from linguistics to contributing to our understanding
of semantics.

In the meantime, robots and conversational agents will continue to be deployed.
It is critical that we have confidence in our systems and that input merely
being out of the training set does not cause catastrophic failure.
We demonstrate one step toward achieving this goal: a principled way to enable
networks to generalize out of the training set.
Many open problems remain, key among them: is there a way to convert a data
augmentation approach into a network architecture that sees through the problem
and generalizes better for a principled reason without the data augmentation.
This would be a powerful tool, which we suspect exists, but have not yet found.

\section{Ethics and broader impacts}

Robots that can competently understand natural language will provide access
to technology for those who need it most: those who have physical limitations,
those with limited access to education, etc. This can have tremendous positive
impact as well as negative consequences. For example, such robots may displace
human workers leading to widespread job loss. We already see this in that
bots are taking over many interactions that would otherwise have gone through
a customer support representative. The future impact of language-driven robots
and conversational agents will depend on a combination of researchers who
tailor systems to augment rather than displace workers as well as politicians who
create safety nets and training for displaced workers.

Our adoption of methods which attempt to be transparent, i.e., by forcing 
the models to reason in the open through attention maps, can help with
pinpointing errors. Currently, complex systems, end-to-end models in particular, 
have an attribution problem. One is largely uncertain about why they fail. A robot that harms someone, one that
discriminates overtly or covertly, etc. should be designed in such a way that
one can determine why these actions were taken, to assign financial and legal liability
as we do with all other engineered systems.


\section{Acknowledgments}
This work was supported by the Center for Brains, Minds and Machines, NSF STC award 1231216, 
the MIT CSAIL Systems that Learn Initiative, the CBMM-Siemens Graduate Fellowship,
the DARPA Artificial Social Intelligence for Successful Teams (ASIST) program,
the United States Air Force Research Laboratory 
and United States Air Force Artificial Intelligence Accelerator under Cooperative Agreement Number FA8750-19-2-1000, and the Office of Naval Research under Award Number N00014-20-1-2589 and Award Number N00014-20-1-2643.
The views and conclusions contained in this document are those of the authors and should not be interpreted as representing the official policies, either expressed or implied, of the U.S. Government. The U.S. Government is authorized to reproduce and distribute reprints for Government purposes notwithstanding any copyright notation herein.

\bibliography{references}  
\bibliographystyle{acl_natbib}

\end{document}

%% file: preamble.tex
\usepackage{times} 
\usepackage{amsmath} 
\usepackage{amssymb}  
\usepackage{algorithm}
\usepackage[noend]{algpseudocode}
\usepackage{comment}
\usepackage{graphicx}
\usepackage{subfigure}
\usepackage{tikz}
\usepackage{etoolbox}
\usepackage{xspace}
\usepackage{xcolor}
\usepackage{comment}
\usepackage[utf8]{inputenc}
\usepackage[T2A,T1]{fontenc}
\usepackage{booktabs}
\usepackage{paralist}
\usepackage{cleveref}
\usepackage[export]{adjustbox}
\usepackage{scalerel}
\usepackage[space]{grffile}
\usepackage{siunitx}
\graphicspath{{images/}}

\definecolor{Red}{rgb}{1,0,0}
\definecolor{Green}{rgb}{0,0.69,0}
\definecolor{Blue}{rgb}{0,0,1}
\definecolor{LightBlue}{rgb}{0,0.5,1}
\definecolor{veryLightBlue}{rgb}{0.85,0.98,1}
\definecolor{veryLightGreen}{rgb}{0.6,1,0.6}
\definecolor{Skin}{rgb}{1,0.71,0.69}
\definecolor{Grey}{rgb}{0.5,0.5,0.5}
\definecolor{LightGrey}{rgb}{0.6,0.6,0.6}
\definecolor{Black}{rgb}{0,0,0}
\definecolor{White}{rgb}{1,1,1}
\newcommand{\red}{\color{Red}}
\newcommand{\green}{\color{Green}}
\newcommand{\blue}{\color{Blue}}

\newcommand{\grey}{\color{Grey}}

\newcommand{\orange}{\color{orange}}

\newbool{blind}
\setbool{blind}{false}

\ifbool{blind}
{\AtBeginEnvironment{blind}{\comment}%
  \AtEndEnvironment{blind}{\endcomment}}{}

\makeatletter
\DeclareRobustCommand\onedot{\futurelet\@let@token\@onedot}
\def\@onedot{\ifx\@let@token.\else.\null\fi\xspace}

\def\etal{\emph{et al}\onedot}
\newcommand{\eg}{e.g.,\xspace}

\newcommand{\ie}{i.e.,\xspace}

\makeatother

\usetikzlibrary{calc}
\usetikzlibrary{fit}
\usetikzlibrary{positioning}
\usetikzlibrary{decorations.markings}
\usetikzlibrary{shapes.geometric}
\usetikzlibrary{shapes.multipart}
\usetikzlibrary{bayesnet}
\usetikzlibrary{tikzmark}
\usetikzlibrary{backgrounds} 
\usetikzlibrary{shapes}
\usetikzlibrary{3d}
\usetikzlibrary{arrows.meta}
\usetikzlibrary{shapes.geometric}

\definecolor{pinegreen}{cmyk}{0.92,0,0.59,0.25}
\definecolor{royalblue}{cmyk}{1,0.50,0,0}
\tikzstyle{cblue}=[circle, draw, thin,fill=cyan!20, scale=0.8]
\tikzstyle{obs}=[circle, draw, thin,fill=gray!20, scale=0.8]
\tikzstyle{qgre}=[circle, draw, scale=0.8]
\tikzstyle{rpath}=[thick, black, opacity=0.4]

\tikzstyle{pathnode}=[circle, draw, scale=0.04]
\tikzstyle{graphnode}=[circle, draw, scale=0.15,ultra thin]
\tikzstyle{graphedge}=[-{Latex[black,length=0.15ex,width=0.15ex]}, shorten >= 0.04ex, ultra thin]
\tikzstyle{graphdashed}=[dash pattern=on 0.2pt off 0.1pt]